# Dynamic Model Agnostic Reliability Evaluation of Machine-Learning Methods Integrated in Instrumentation & Control Systems

**Edward Chen[1*], Han Bao[2], and Nam Dinh[1]**

[1]North Carolina State University, Raleigh, NC
[2]Idaho National Laboratory, Idaho Falls, ID

## ABSTRACT

In recent years, the field of data-driven neural network-based machine learning (ML) algorithms has grown significantly and spurred research in its applicability to instrumentation and control systems. While they are promising in operational contexts, the trustworthiness of such algorithms is not adequately assessed. Failures of ML-integrated systems are poorly understood; the lack of comprehensive risk modeling can degrade the trustworthiness of these systems. In recent reports by the National Institute for Standards and Technology, trustworthiness in ML is a critical barrier to adoption and will play a vital role in intelligent systems' safe and accountable operation. Thus, in this work, we demonstrate a real-time model-agnostic method to evaluate the relative reliability of ML predictions by incorporating out-of-distribution detection on the training dataset. It is well documented that ML algorithms excel at interpolation (or near-interpolation) tasks but significantly degrade at extrapolation. This occurs when new samples are "far" from training samples. The method, referred to as the Laplacian distributed decay for reliability (LADDR), determines the difference between the operational and training datasets, which is used to calculate a prediction's relative reliability. LADDR is demonstrated on a feedforward neural network-based model used to predict safety significant factors during different loss-of-flow transients. LADDR is intended as a "data supervisor" and determines the appropriateness of well-trained ML models in the context of operational conditions. Ultimately, LADDR illustrates how training data can be used as evidence to support the trustworthiness of ML predictions when utilized for conventional interpolation tasks.

## 1. INTRODUCTION

Integration of machine learning (ML) into instrumentation and control systems has grown in interest in recent years. ML systems have been used for enhanced plant diagnostics [1], automated scheduling of maintenance tasks [2], autonomous control [3], etc. Typically for these ML models, a training dataset defines model function by learning a correlation between input and target values. Unlike conventional software programs, where function can be described succinctly by the implemented algorithm, the function achieved through training is governed by the multiplication of nondescript weights and biases. The functional correctness of the model thus cannot be comprehensively interpreted and verified by stakeholders of the system. This issue is significant as numerous examples of ML models failing due to various hidden root causes and failure mechanisms such as regressional inconsistencies [4], inherent distributional rigidity [5], metric optimization failures [6], or unintended adversarial examples [7]. While validation and verification of these ML models has grown, these models still experience significant drops in performance when applied to real-world operational conditions [8]. In reports by the National Institute for Standards and Technology (NIST) [9] and the United States Nuclear Regulatory Commission [10], they indicate that trustworthiness in ML is a critical barrier to adoption and will play a vital role in the safe, accountable, and secure operation of intelligent systems.

---

* echen2@ncsu.edu

Fundamentally, it is well known that ML models excel at interpolation (or near-interpolation) tasks and experience performance degradation at extrapolation. One possibility to develop trustworthiness in ML predictions is by assessing how well the real-world operational data matches the training data of the model. If the operational and training data are similar (i.e., interpolation), we can assume there is some basis for trust in the model's predictions. This is analogous to inductive reasoning, where the samples in the training database act as specific evidence to support generalized conclusions made by the ML model. However, determining when (or "how far") a new sample is considered an extrapolation task is a challenge and is the basis for out-of-distribution (OOD) [11] detection research.

Thus, the purpose of this work is to determine "how far" a new sample must be to be considered OOD and how this value can be used to support trust in ML integrated systems. We established a method to evaluate the reliability of model predictions relative to the training data. The fundamental assumption being that training data can be scrutinized for correctness and representative of the mission objective. In this work, training data is analogous to student education and serves as the knowledge base. An out-of-distribution detection method is applied on the operational data to determine reliability. The method, named Laplacian distributed decay for reliability (LADDR), can be used to determine the relative reliability of model predictions in real-time and can be applied to any data-driven time-invariant (or memoryless) model.

In section 2, the theoretical background behind trustworthiness in ML systems is discussed and the concept of relative reliability is presented. In section 3, the mathematical details within LADDR are discussed and a method for parameter optimization is presented. In section 4, a case study using LADDR within an ML integrated instrumentation system is presented based on the Nearly Autonomous Management and Control (NAMAC) [1, 3] system. Finally, in section 5, a brief discussion on the results and conclusion of this work is presented.

## 2. THEORETICAL BACKGROUND

### 2.1 Trustworthiness

The concept of trustworthiness is a human conceived aspect and is relative based on the contextual scenario of the application. Simply applying quantitative numbers and verifiable facts does not necessarily establish trustworthiness of a system by humans. Therefore, building a foundation of trustworthiness in ML integrated systems is a convoluted task. Currently, one of the most concise and up-to-date definitions of trustworthiness as a human aspect can be found in [9]. In their work, seven characteristics of trustworthiness are described (Figure 1), each embodying a socio-technical attribute related to the activities of the development and deployment of an ML model.

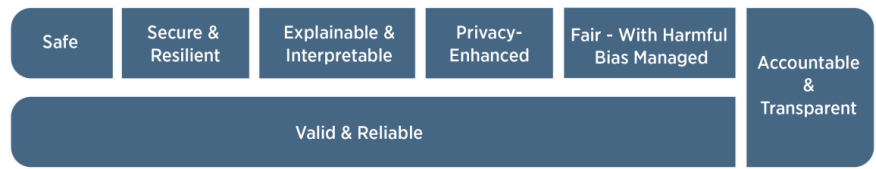

**Figure 1. Characteristics of trustworthy ML models [9].**

In this work, we focus on the Valid & Reliable category, specifically, the reliability subcategory. In this subcategory, reliability is defined as "the ability of an item to perform as required, without failure, for a given time interval, under given conditions [9]". Unfortunately, this definition is difficult to apply to ML integrated systems for a variety of reasons.

### 2.2 ML Reliability

The genesis of reliability was based on analog systems, where components could fail due to mechanical degradation. Reliability assumes that a component has a limited life span due to environmental degradation (e.g., corrosion) such that the time to functional failure (or inverse reliability) is probabilistic in nature. That is not the case for software which includes ML models. While hardware components (e.g., integrated circuits) that support the software can fail and be modeled, software reliability is not the same case. Conventional methods for functional validation in development, and even pre-deployment, phases are not sufficient to guarantee reliability in the deployment phase [12]. While functional testing can improve the quality of the software, it cannot guarantee the failure-free operation of the software. Inconsistencies in design, errant assumptions, and incorrect implementations can lead to hidden latent defects (like landmines) in the source code. Thus, in [13, 14, 15], software failures are viewed as systematic failures of the design process, where existing statistical methods may be inadequate to determine reliability. If the same set of input conditions are consistently provided to software, the output will remain correct regardless of the number of tests performed. Unless a specific sequence and combination of unknowable conditions are supplied, the system will never fail (see [16] for an example of unknowable conditions). Apparently "reliable" software can thus unexpectedly fail at any time. Otherwise, the defect will remain hidden and have no visible impact on the overall system function. The concept of degradation over time on software also does not apply as software is virtual and without physical form. The generic hardware required to run software can be easily swapped making the theoretical life infinite until irrelevant. Therefore, the conventional definition of reliability for software is problematic.

Unfortunately, the current definition of reliability is also used for ML integrated systems and experiences similar challenges. Take for example, a non-reinforcement memoryless deep feedforward neural network (FNN), which approximates an unknown correlation through array multiplication of neuron layers. Typically, training samples ($x \in X$) are used tune the correlation through backpropagation of weights and biases. During training, the model is validated and verified on withheld samples within $X$ which is assumed to be sampled from the operational set ($W$). $W$ describes the set of all inputs and outputs in the realistic application of the system. Suppose during operation, the model is supplied with inputs, $x' \in W$, that mimic training samples from $X$ such that $x_1' = x \in X$. As the model was validated on the training set, it is guaranteed it will perform as expected for any time interval and thus achieve perfect reliability. However, realistically, the training set will always be a subset of the operational set (i.e., $X \subset W$) due to limitations in dataset construction. This suggests that the domain of $x'$ is larger than $X$ where samples can be drawn from $W$ that are not in $X$ as observed in Figure 2. FNNs are known to have degraded performance at extrapolation, or generally when $x_2' \notin X$.

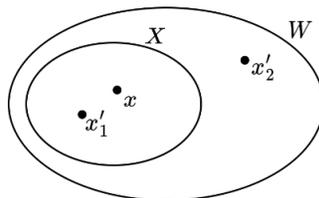

**Figure 2. Samples drawn from the training set, as well as the operational set.**

A rudimentary definition of reliability in ML can be described as the probability that new samples are drawn exclusively from $X$ and not within $W\backslash X$. This is the basis for OOD detection [11] which addresses the question, at what point is a new sample within $X$ and not within $W\backslash X$. To refine the definition of reliability, we use an arbitrary two feature dataset as an example. In Figure 3(a), the training set contains a single point, $x$, and four sample points (i.e., $a$ through $d$). At what distance is point $a$ considered to be too far from the $x$ and thus OOD? Conversely, point $d$ is the closest to $x$, therefore is it the most in-distribution (ID)? Similarly, in Figure 3(b), point $e$ is internally far from training points while point f is externally far. The Euclidean or Mahalanobis distance [17] are two functions used to calculate separation (i.e., $\varepsilon_1$ and $\varepsilon_2$) but it is difficult (and subjective) to determine a consistent decision boundary for when $\varepsilon$ is large enough to be considered OOD. It also overlooks the fact that datapoints may be internally far versus externally far or in densely versus sparsely populated clusters.

In this work, we propose the concept of relative reliability of ML predictions as to whether samples points can be represented by a limited set of local training points. Mathematically, if $x_1' = x \in X$, then the predictions are considered reliable (i.e., $R = 1$) as they emulate training data. If $x_1' \neq x$ but $x_1' \in X$ then the predictions are somewhat reliable (i.e., $0 < R < 1$) based on the function, $L(\vec{x}_1', \vec{x}, q)$, between $x_1'$ and the $x \in X$. The function $L(\vec{x}_1', \vec{x}, q)$ is defined by a localized double exponent decay function (also known as a Laplacian distribution) applied to all points $x \in X$. For the example shown in Figure 3(a), the relative reliability of a prediction at any point can be mapped as the contour plot shown in Figure 4(a). As input samples $a$ through $d$ move further away from the training point $x$, the reliability of the prediction decays exponentially. In Figure 4(b), the superposition of Laplacian distributions over all training points illustrates how internal samples (i.e., point $e$) may have higher reliability than the external sample (i.e., point $f$), but may still be OOD relative to the training data (i.e., low reliability). The contour plots in Figure 4 are known as reliability maps. This method of reliability determination can be scaled to any number of features and include the target of each sample as well. The method proposed is called LADDR.

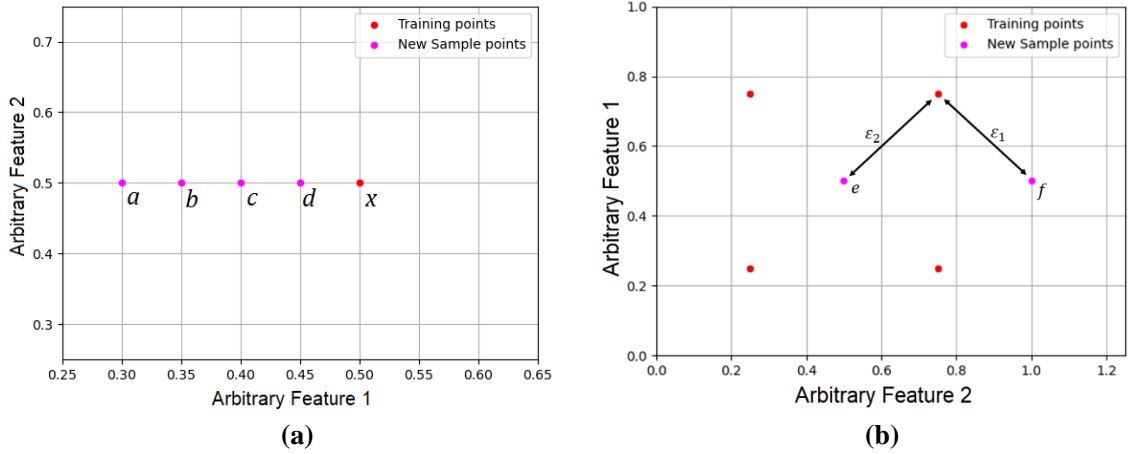

**Figure 3.** (a) Single training point. (b) Multiple training points.

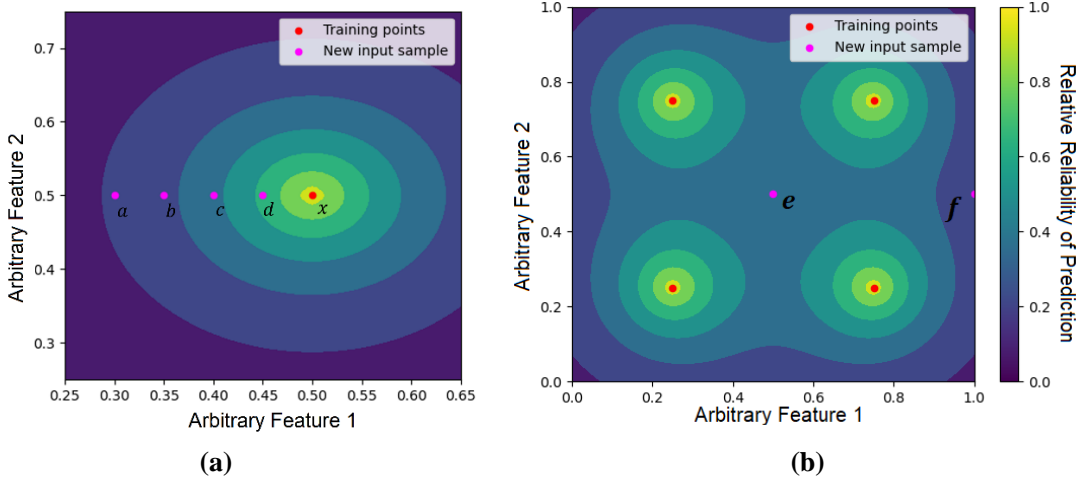

**Figure 4.** Relative reliability of model prediction over (a) a single point and (b) multiple points.

## 3. METHODOLOGY

### 3.1 Formulation of LADDR

LADDR is adapted from the 1D normalized Laplacian distribution from [18] (Equation (1)).

$$L(\vec{x}', \vec{x}, V) = \frac{1}{2\alpha} \exp\left(-\frac{|x - \mu|}{\alpha}\right) = \exp(-2D(\vec{x}', \vec{x})) \qquad (1)$$

$$D(\vec{x}', \vec{x}) = \sqrt{(\vec{x}' - \vec{x})^T V^{-1} (\vec{x}' - \vec{x})} \tag{2}$$

Here, $\vec{x}'$ is the new sample vector, $\vec{x}$ is the nearest training sample, $D(\vec{x}', \vec{x}, V)$ is the Mahalanobis distance, $L(\vec{x}', \vec{x}, V)$ is the relative reliability from $\vec{x}'$ to $\vec{x}$, and $V$ is the covariance structure matrix for distribution spread in n-dimensions, as in Equation (3). For the desired behavior where $L(\vec{x}', \vec{x}, V) = 1$ when $x'_1 = x \in X$, the scale factor is set to $\alpha = 0.5$. $V$ is a diagonal matrix where $\beta_n$ represents the decay rate in the axis of each feature. It is not the covariance of the dataset. For a single training sample, $\tilde{L}(\vec{x}', \vec{x}, V)$, will produce a number between 0 and 1, where 1 represents perfect alignment of the training and the input sample.

$$V = \begin{pmatrix} \beta_1 & \cdots & 0 \\ \vdots & \ddots & \vdots \\ 0 & \cdots & \beta_n \end{pmatrix} \tag{3}$$

A superposition of Equation (1) was used to generate the relative reliability maps seen in Figure 4 and is the relative reliability of model predictions to the training data knowledge base.

## 3.2 Optimization of LADDR Parameters

Without a doubt, the hyper parameters within LADDR will influence the outcome of the model. Therefore, in this section we provide guidance and analytical metrics to optimize LADDR. First, to facilitate optimization, we introduce the concept of extrapolation diameter, $\gamma$, which represents how far a single training point can be 'stretched' to cover interpolation tasks. The extrapolation diameter is defined per input feature as the width at $L(\cdot) = 20\%$. In Figure 5(a), for instance, a training point for feature 1 is located at $x = 0.5$ with a specified $\gamma = 0.36$. The extrapolation radius (or half the diameter) $\gamma_{1/2} = 0.18$ suggests that samples less than $x = 0.68$ or greater than $x = 0.32$ are a fifth as reliable as closer samples to $x = 0.5$. For users, this allows them to specify the literal distance when data points become OOD. This also allows hyperparameters to be optimized relative to the extrapolation diameter which is more interpretable than the covariance structure matrix. By specifying the extrapolation diameter for each feature, the covariance structure matrix can be solved for using Equation (4). In Figure 5(b), the variance in arbitrary feature 2 is set to five times greater than arbitrary feature 1.

$$V = \frac{\vec{\gamma}_{\frac{1}{2}}^T \vec{\gamma}_{\frac{1}{2}}}{(-\ln(L(\cdot))^2} \tag{4}$$

Instances where $\beta_1 \neq \beta_2$ in $V$ are evident when the importance of training features are not equivalent. Selection of $\beta$ parameters is relative to the optimization goal and the usage context of the system. In Figure 6, a utilization framework is presented where LADDR is integrated with an arbitrary predictive ML model. Within this framework, the training data used to train the model serves as the knowledge base for LADDR. For a new sample, if LADDR determines the sample and subsequent prediction are likely to be in-distribution, the prediction is accepted, and the system can proceed normally. Alternatively, if it is likely to be OOD, the prediction is rejected regardless of the actual correctness or capability of the model, and auxiliary functions are engaged.

If stakeholders are concerned about safety, LADDR may be configured where only samples that emulate training data are accepted, whereas all others are rejected. Conversely, LADDR can be configured to maximize model performance while only rejecting significantly different contextual scenarios. In this respect, LADDR acts as a 'supervisor', filtering out scenarios that are irrelevant to the ML model. As LADDR acts as a filter, it is expected the ML model's performance will degrade whenever LADDR rejects a correct prediction. However, if LADDR were to accept all predictions, it is likely an incorrect prediction is not filtered and is a peril to the operational safety of the system. These two perspectives are used to form two analytical metrics known as degradation ($f_D$) and peril ($f_P$) seen in Equations (5) and (6), respectively. These metrics are used to minimize what is most important to stakeholders of the ML system in the context of their application.

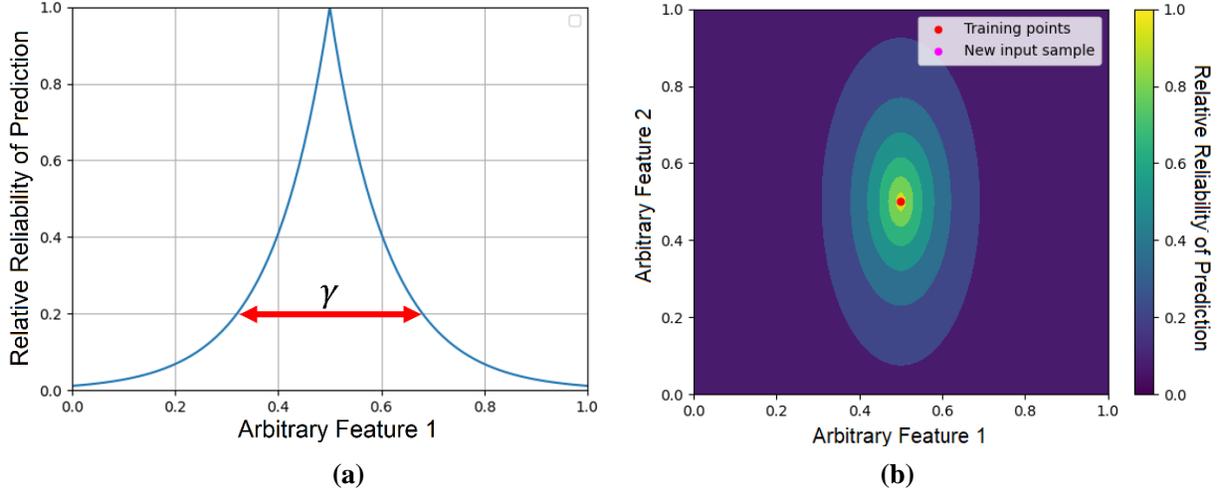

**Figure 5. (a) Extrapolation diameter. (b) Elliptical distribution generated by $\beta_2 = 5\beta_1$ in $V$.**

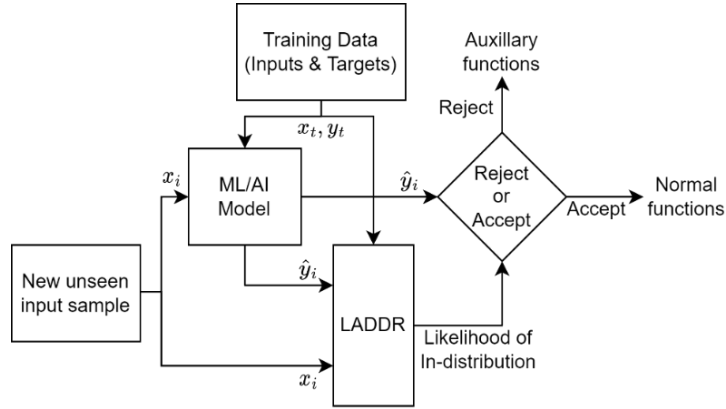

**Figure 6. Basic framework of LADDR coupled with an ML control or instrumentation system.**

$$f_D = \frac{R_o}{R_o + A_o} \tag{5}$$

$$f_P = \frac{A_x}{A_x + A_o} \tag{6}$$

Here, $R_o$ and $A_o$ are the total number of rejected and accepted predictions that were correct while $R_x$ and $A_x$ are the total number of rejected and accepted predictions that were incorrect. The degradation metric, Equation (5), describes the ratio of correct predictions made by the model but were subsequently rejected over the total number of correct predictions that were rejected or accepted. A value of 0 suggests that all correct predictions made by the model were also accepted by LADDR while a value of 1 suggests that all correct predictions were rejected. This can occur if the extrapolation diameter specified for each feature in LADDR is too small such that only the immediate vicinity of the training data points is considered sufficient evidence to justify predictions. For stakeholders concerned about peak performance and less so about functional safety of the system, the extrapolation diameters can be chosen to minimize this parameter.

Alternatively, if the system is considered safety-critical, the extrapolation diameters are chosen to minimize the peril metric, Equation (6). This metric describes the ratio of incorrect predictions accepted over the total number of predictions accepted by LADDR. A value of 0 is where all accepted predictions are correct while a value of 1 is where that all accepted predictions are incorrect. Correctness is determined relative to the specified acceptance criteria by stakeholders (e.g., $\pm 1\%$). Peril ($f_P$) grows as extrapolation diameters are increased and represent when limited data is used to make gross predictions.

However, importantly, if degradation is minimized, peril will grow, and vice versa, as the extrapolation diameters that minimize either are different. Therefore, we also introduce a third metric, ineptitude ($f_f$), which combines of the degradation and peril metrics (Equation (7)). The ineptitude metric describes the total number of rejected correct predictions and accepted incorrect predictions over the total number of predictions that are filtered. A value of 1 suggests that all predictions filtered are erroneous while a value of 0 suggests LADDR is perfectly proficient at accepting correct—but rejecting incorrect—predictions. This metric can be minimized if both performance and safety are desired by stakeholders.

$$f_f = \frac{R_o + A_x}{A_o + R_x + R_o + A_x} \tag{7}$$

## 4. CASE STUDY

The LADDR framework presented in Figure 6 is applied to a diagnostic digital twin (DT) model [1] developed for the Nearly Autonomous Management and Control (NAMAC) [3] system. The DTs are deep FNN developed for partial loss-of-flow accident scenarios to predict the peak fuel centerline temperature ($T_{FCL}$). The datasets were gathered and simulated using a GOTHIC™ model of the Experimental Breeder Reactor-II (EBR-II) [19]. In EBR-II, two separate primary sodium pumps, designated as P1 and P2, provide coolant flow through the core. In the postulated scenario, P1 partially loses pump rotational speed, thus decreasing the overall coolant flow through the core block. The scenario is monitored by the diagnostic DT. Based on NAMAC's automated recommendations, the rotational speed of P2 is increased to compensate. The scope of the numerical demonstration can be represented by the time-dependent curve of the rotational speed of P1 Equation (8).

$$\omega_1(t) = \omega_0 \left(1 - \frac{1 - (\omega_1)_{end}}{T_1} t \right), \quad t_0 \leq t \leq t_0 + T_1 \tag{8}$$

Here, $\omega_0$ is the nominal pump speed, $T_1$ is the ramp-down duration, $(\omega_1)_{end}$ is the normalized P1 end speed, and $t_0$ is the transient start time. By varying pump end speed and ramp down duration, different profiles can be simulated. Core flow rate, plenum temperatures, and $T_{FCL}$ are sampled for 2000 time-steps per transient (Table 1). In column 1, the reference label is provided; column 2 shows the number of transients per scenario; column 3-4 shows the degree of degradation; and columns 5-6 shows percentage of data used for training and forming LADDR knowledge base vs. used in testing. In Figure 7, the transient data is plotted. Each color represents a different transient from start to end. The set D2 was not used to train the DT and is used to gauge the performance of LADDR when an entirely different scenario is provided. The objective of the diagnostic DT is to predict $T_{FCL}$ within $\pm 10°C$ of the simulated temperature and serves as the acceptance criteria for predictions. As $T_{FCL}$ is generally an unobservable variable due to obstructions within the fuel bundle, establishing the reliability of model predictions is critical to operational safety. For this case study, the upper plenum temperature and total core flow rate are used as inputs to train the DTs while $T_{FCL}$ is used as the predictive target. The same three parameters are used as the knowledge base within LADDR. Finally, note that the post-training performance of the diagnostic DT in this case study is irrelevant. Model performance is intentionally poor to demonstrate the capability of the LADDR framework at rejecting and accepting predictions based on training data knowledge.

Table 1. Size of training and testing sets used within LADDR and for the FNN DTs.

| Label | # of Episodes | $T_1$ | $(w_1)_{end}$ (% of $w_0$) | % Used in Training | % Used in Testing |
|---|---|---|---|---|---|
| D1 | 1024 | 467.81 | 51.6 – 100 | 10% | 90% |
| D2 | 250 | 467.81 | 0 – 38.7 | Not Used | 100% |

In Figure 8a, the reliability map is generated for Figure 7a where only the inputs features are considered for reliability. As the inputs are 2-dimensional, the reliability map illustrates where there is a high degree of training evidence to support model predictions. In Figure 8b, the simulated $T_{FCL}$ is also considered as evidence and a 2-dimensional slice of the 3-dimensional reliability map for when $T_{FCL} = 0.877$ is shown. Figure 8b illustrates that even if input data is similar to training data, if the prediction is not correct, there is very little reliability in the outcome.

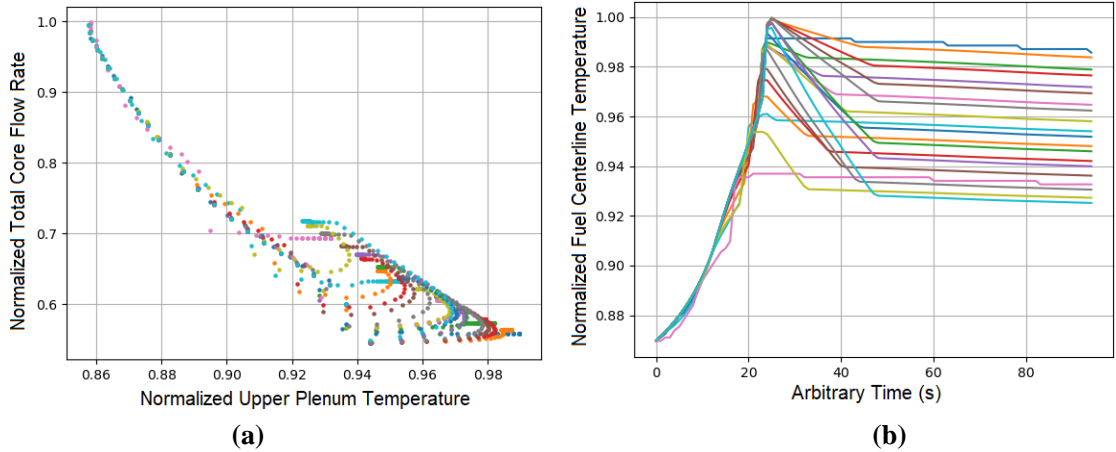

Figure 7. (a) Normalized input training data. (b) Normalized fuel centerline temperature transients.

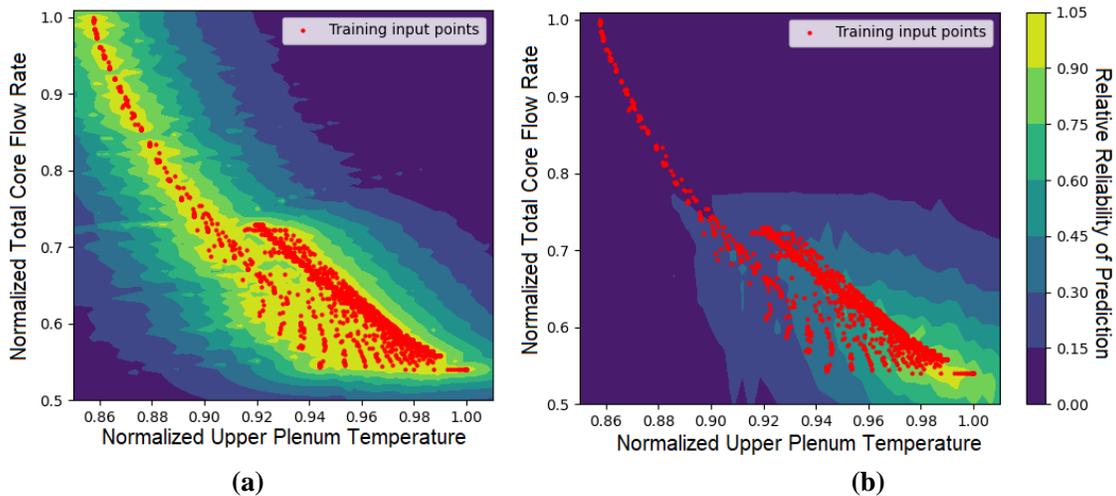

Figure 8. a) Reliability map for inputs. b) Reliability map for inputs for normalized simulated $T_{FCL} = 0.877$.

In Figure 9a, the DT prediction for $T_{FCL}$ on a single loss-of-flow transient from set D1 is shown. In Figure 9b, a prediction on set D2 is shown. The predictive capability of the model is intentionally poor (dashed red line) and makes several errors when compared against the simulated $T_{FCL}$ (dotted red line). In the areas of poor predictions, the reliability of the prediction (blue solid line) drops precipitously. In these areas, the model predictions should be rejected as they lack sufficient supporting training data evidence. An arbitrary reject/accept boundary is established at 0.5 to examine the proposed analytical metrics developed for the framework in Figure 6. In Table 2, the optimized extrapolation diameters based on stakeholder concerns for the training set within D1. In column one, the metric minimized is specified, in columns 2-4, the optimized extrapolation diameters are shown when the metric is minimized, and in columns 5-7, the analytical metrics are shown.

Table 2. Analytical metrics for extrapolation diameter optimization.

| Minimized Metric | Optimized Normalized Extrapolation Diameters | | | Analytical Metrics | | |
|---|---|---|---|---|---|---|
| | UP Temp. | Total Core Flow Rate | $T_{FCL}$ | Peril | Degradation | Ineptitude |
| Peril | 0.0254 | 0.0254 | 0.064 | 0 | 0.105 | 0.085 |
| Degradation | 0.0254 | 0.0254 | 0.113 | 0.116 | 0 | 0.105 |
| Ineptitude | 0.0254 | 0.0254 | 0.072 | 0.027 | 0.079 | 0.084 |

In Table 3, the total number of incorrect and correct rejections and acceptances are shown for the testing sets D1 and D2. Although the extrapolation diameters are minimized for the training set D1, LADDR can still perform well on set D2, filtering out a majority of unsafe predictions. The degradation is expected to be higher for set D2 as the scenarios are significantly different. However, as D2 is not part of LADDR's knowledge base, the higher rejection rate is associated with lack of training data evidence to support model predictions even if they are within the $\pm 10^{o}C$ acceptable margin.

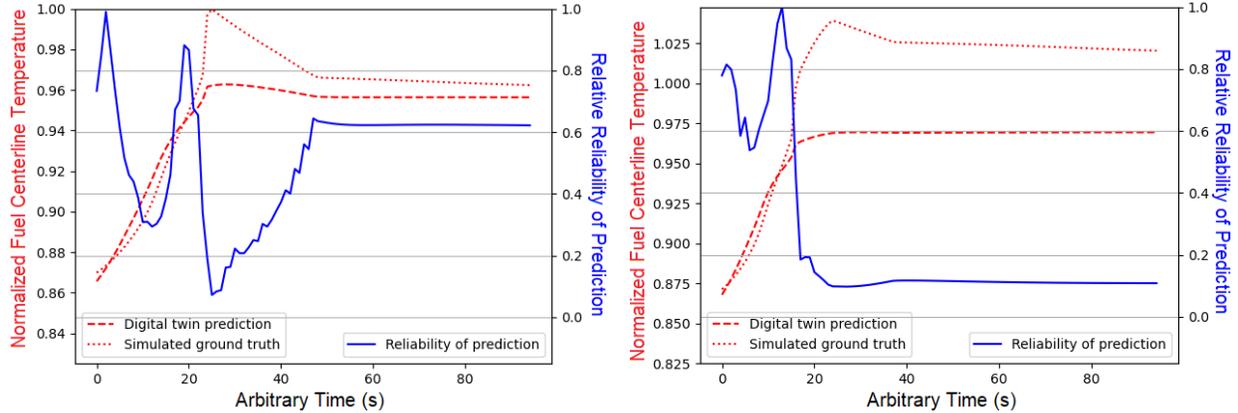

**Figure 9. Prediction reliability (blue line) against model prediction and simulated truth (red dotted line).**

**Table 3. Summary of LADDR performance of the developed loss-of-flow transient datasets.**

| Dataset | Correct Accept | Correct Reject | Incorrect Accept | Incorrect Reject | Peril | Degradation | Ineptitude |
|---|---|---|---|---|---|---|---|
| D1 | 13,825 | 4,084 | 635 | 931 | 4.4% | 6.3% | 8.04% |
| D2 | 6,865 | 3,487 | 400 | 3,783 | 5.5% | 35.5% | 28.8% |

## 5. CONCLUSION

In this work, a novel approach called Laplacian distributed decay for reliability (LADDR), was presented, and used to determine the relative reliability of ML predictions from the perspective of the training data evidence in the construction of the models. The premise being that, regardless of if the model's predictions are correct or incorrect, unless sufficient training data exists to support those conclusions, the predictions must be rejected as they represent out-of-distribution tasks. An ML model is paired with LADDR as a filter was demonstrated on a diagnostic digital twin [1] for fuel centerline temperature prediction within the Nearly Autonomous Management and Control system [3]. The preliminary results demonstrate the utility of LADDR at filtering out predictions that are unreliable relative to the training database. Three additional optimization metrics were also presented to allow optimization of LADDR parameters based on stakeholder concerns. Peril, which dictates the rate of incorrect predictions not filtered; degradation, which dictates the percentage of performance drop as a result of LADDR acting as a filter function to the ML model; and ineptitude, which is a culmination of the other two metrics. Currently, LADDR has only been tested on 3D training sets. As most training sets can contain upwards of 10 features, we plan to verify the capability of LADDR on higher dimensions and improve upon the optimization techniques for the extrapolation diameters. It is anticipated that LADDR can improve the trustworthiness of ML models by acting as a "supervisor" for predictions made by intelligent systems.

## ACKNOWLEDGMENTS

This project was authored by a contractor of the U.S. Government under U.S. Department of Energy Contract No. DE-AC07-05ID14517.